\documentclass[sn-mathphys,Numbered]{sn-jnl}

\usepackage{graphicx}%
\usepackage{multirow}%
\usepackage{amsmath,amssymb,amsfonts}%
\usepackage{amsthm}%
\usepackage{mathrsfs}%
\usepackage[title]{appendix}%
\usepackage{xcolor}%
\usepackage{textcomp}%
\usepackage{adjustbox}
\usepackage{manyfoot}%
\usepackage{booktabs}%
\usepackage{algorithm}%
\usepackage{algorithmicx}%
\usepackage{algpseudocode}%
\usepackage{listings}%
\usepackage{array}
\usepackage{rotating}
\usepackage{tabularx}
\emergencystretch=\maxdimen
\usepackage{makecell}
\usepackage{xcolor}

\theoremstyle{thmstyleone}%
\theoremstyle{thmstyletwo}%

\theoremstyle{thmstylethree}%

\begin{document}

\title[Mask Focal Loss: A unifying framework for dense crowd counting with canonical object detection networks]{Mask Focal Loss: A unifying framework for dense crowd counting with canonical object detection networks}


\author[1]{\fnm{Xiaopin} \sur{Zhong}}\email{xzhong@szu.edu.cn}

\author[2]{\fnm{Guankun} \sur{Wang}}\email{gkwang@link.cuhk.edu.hk}

\author*[1]{\fnm{Weixiang} \sur{Liu}}\email{wxliu@szu.edu.cn}

\author[1]{\fnm{Zongze} \sur{Wu}}\email{zzwu@szu.edu.cn}

\author[3]{\fnm{Yuanlong} \sur{Deng}}\email{dengyl@szu.edu.cn}

\affil*[1]{\orgdiv{College of Mechatronics and Engineering}, \orgname{Shenzhen University}, \orgaddress{\street{Nanshan}, \city{Shenzhen}, \postcode{518060}, \state{Guangdong}, \country{China}}}

\affil[2]{\orgdiv{Department of Electronic Engineering}, \orgname{The Chinese University of Hong Kong}, \orgaddress{\street{Shatin}, \city{N.T.}, \postcode{999077}, \state{Hong Kong}, \country{China}}}

\affil[3]{\orgname{Shenzhen Institute of Technology}, \orgaddress{\street{Pingshan}, \city{Shenzhen}, \postcode{518060}, \state{Guangdong}, \country{China}}}


\abstract{As a fundamental computer vision task, crowd counting plays an important role in public safety. Currently, deep learning based head detection is a promising method for crowd counting. However, the highly concerned object detection networks cannot be well applied to this problem for three reasons: (1) Existing loss functions fail to address sample imbalance in highly dense and complex scenes; (2) Canonical object detectors lack spatial coherence in loss calculation, disregarding the relationship between object location and background region; (3) Most of the head detection datasets are only annotated with the center points, i.e. without bounding boxes. To overcome these issues, we propose a novel Mask Focal Loss (MFL) based on heatmap via the Gaussian kernel. MFL provides a unifying framework for the loss functions based on both heatmap and binary feature map ground truths. Additionally, we introduce GTA\_Head, a synthetic dataset with comprehensive annotations, for evaluation and comparison. Extensive experimental results demonstrate the superior performance of our MFL across various detectors and datasets, and it can reduce MAE and RMSE by up to 47.03\% and 61.99\%, respectively. Therefore, our work presents a strong foundation for advancing crowd counting methods based on density estimation.}


\maketitle

\section{Introduction}\label{sec1}

Crowd counting has always been the hot topic in public safety. Its goal is to estimate the number of pedestrians in different secnerios, which plays an important role in risk perception and early warning, traffic control and scene statistical analysis. In public spaces such as subways, squares, shopping malls and sports events, if the number and distribution of pedestrians can be calculated in real time, the staff can be more targeted for statistical analysis, order maintenance and dangerous  prevention~\cite{sindagi2018survey, wang2021improved, gao2020cnn, fekri2023developing}. As a consequence, the work in this field is valuable and has attracted a high degree of attention. In the past few years, computer vision based methods become the most important means of crowd counting in public places because of their advantages such as long-distance detection, passive sensing and simultaneous perception of multiple targets. However, accurate counting is still a challenging unsolved problem due to scene complexity, large-scale variation and high crowd density in the real world. 

\begin{figure*}[!t]
\centering
\includegraphics[width=5in]{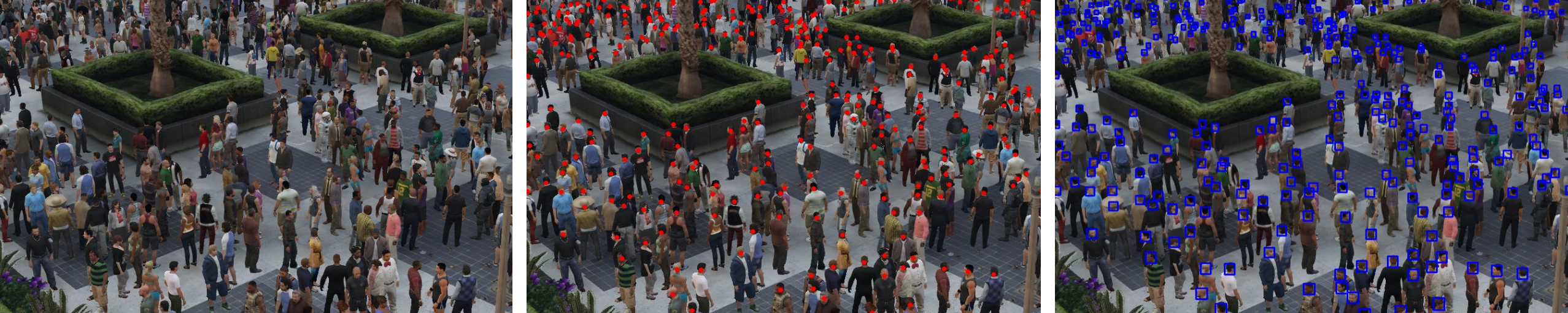}
\caption{An example of high dense and complex crowd counting scene. (left) original image, (middle) image with head center labels, (right) image with head bounding box labels.}
\label{fig1}
\end{figure*}

In the field of crowd counting, there have recently emerged a number of summary works, such as~\cite{sindagi2018survey,tripathi2019convolutional}. Gao et al.~\cite{gao2020cnn} investigated the progress of crowd counting in a single image based on the convolutional neural network (CNN) based density estimation methods in 2020. On the one hand, these literatures do not include the progress in the past two years. On the other hand, they focus only on the methods based on density estimation, ignoring the strategies based on detection for the reason that the state-of-the-art (SOTA) methods on public datasets for crowd counting all use density estimation-based networks without exception. In this paper, we consider mainly detection-based crowd counting with deep learning. Especially in the past two years, deep learning based methods~\cite{gu2023eantrack, gu2022rpformer, yuan2023active, gu2023repformer} have brought a key improvement to object detection based crowd counting. Some researchers~\cite{sam2020locate, song2021rethinking, wang2021dense} replace the whole body detection by detecting pedestrian heads that are difficult to be obscured. Nevertheless, their network structure is specially customized and complex, such as multi-scale and self-attention mechanisms are introduced.  
The above methods and canonical target detection methods still can not outperform the SOTA methods based on density estimation. There are three main reasons for the restriction:
\begin{itemize}

  \item First, the sample imbalance is particularly prominent. Most existing loss functions calculate the positive loss in the entire target area with the same weight. To tackle this, for example in the work of CenterNet~\cite{zhou2019objects}, the distribution of the heatmap ground truth(GT) of the target area attenuates from the center to the borders. However, CenterNet uses a Focal Loss calculation with pixel-level logistic regression which does not take into account the loss contributions according to the distribution change of the heatmap GT. This has weakened the detector's perception of the entire head region, resulting in missing some targets.

  \item Second, in the crowded scene, head regions are confused and interlaced. There is a lot of spatial coherence information between the target area and the background, which is very important for the detector to perceive the targets of interest. However, the object detectors' loss calculation is a hard assignment without taking into account the soft transition from the object location to the background region. This prevents existing loss functions from representing the spatial distribution of GT.

  \item Third, most crowd counting methods are carried out with point-labeled datasets, as shown in Fig.\ref{fig1}(b). At present, there is a lack of dataset with bounding box annotations that supports canonical object detection networks. The point annotation can not be applied to the most mainstream object detection techniques, such as YOLO series, Faster RCNN series and SSD series, which need bounding box annotation. To our knowledge, only four datasets annotated in the way of bounding box can be used for crowd counting, i.e., CroHD~\cite{sundararaman2021tracking}, CrowdX~\cite{hou2022enhancing}, SCUT-Head~\cite{peng2018detecting}, Crowdhuman~\cite{shao2018crowdhuman}, but they are either completely private or partially public or with too low crowd density.
  
\end{itemize}

In order to make the object detection based methods truly return to glory in the field of crowd counting, we first  redivide the positive and negative samples by using a mask to softly distinguish the target region from the background region, so as to maintain the balance between positive and negative samples. We also redefine the weight of the loss contributions according to the heatmap GT and consider the loss contributions at different locations in the target region. The transition from the target location to the background region will no longer be hard, enhancing the perception of the target region and improving the detecting accuracy. Consequently, a so-called Mask Focal Loss is proposed through an ingenious revision of Focal Loss function. Besides, this loss can unify the detectors using different kinds of GTs so that the performance of any method can be truly and fairly compared. Other than that, in order to make the mainstream object detection methods suitable for crowd counting and better evaluate the counting performance, we design and implement a new dataset called GTA\_Head\footnote{https://github.com/gkw0010/GTAV\_Head-dataset.} consisting of 5098 images. We further propose a semi-automatic head box labeling method for large-scale variation, yielding 1732505 head boxes as shown in Fig. \ref{fig1} (right). One of the highlights of our dataset is that our images are derived from the virtual scenes generated by GTAV, a large open-world game, which gives us more freedom to choose the shooting angle, scene and crowd distribution~\cite{wang2019learning}. We hope that this new dataset can provide a better benchmark for future crowd counting research.

Our work will provide a solid foundation for detection based methods to surpass the density estimation-based methods. We apply Mask Focal Loss to several benchmark networks and verify the performance improvement by comparing with traditional losses on multiple datasets.

The rest of this paper is organized as follows. In Section \ref{section:related_works} we investigate the related research work on crowd counting. In Section \ref{section:mask_focal_loss} our proposed Mask Focal Loss is formulated. Experiments and results analysis are provided in Section \ref{section:experiments_results} followed by Section \ref{section:Discussions} in which we further elaborate on the findings from the experimental results. The final section concludes the paper. 

\section{Related works}
\label{section:related_works}

\begin{figure*}[!t]
\centering
\includegraphics[width=5in]{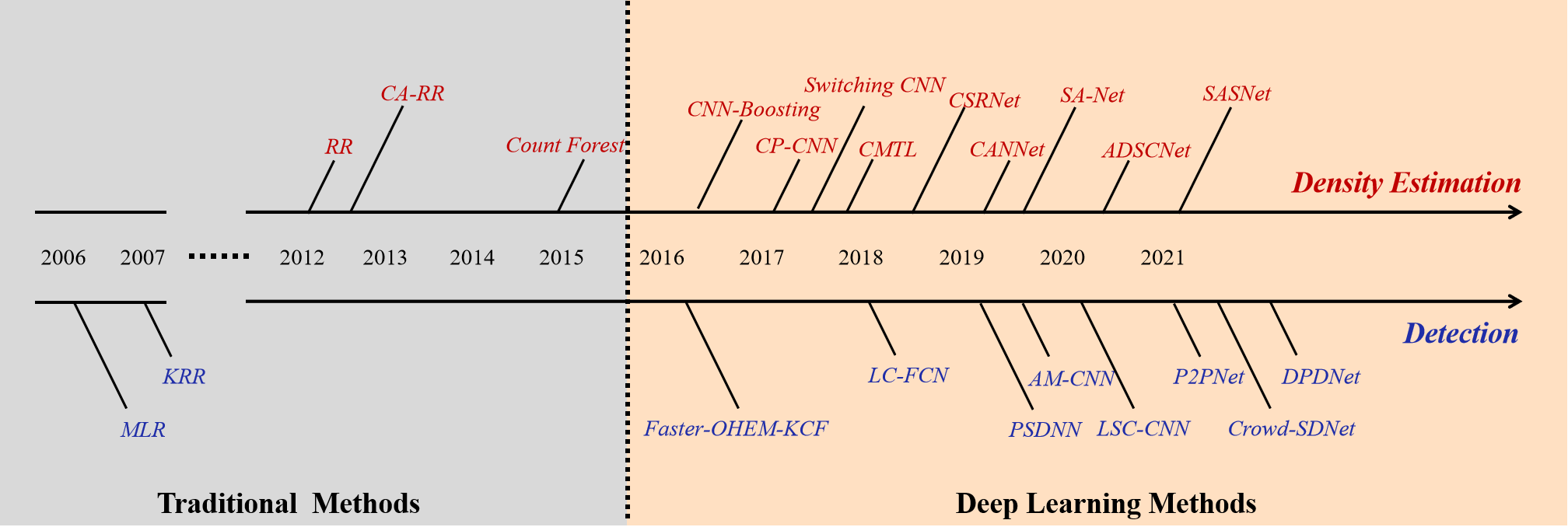}
\caption{Timeline of milestone methods. On the left side of the dotted line are traditional methods, and on the right are methods based on deep learning. Milestone methods: RR~\cite{chen2012feature}, CA-RR~\cite{chen2013cumulative}, Count Forest~\cite{pham2015count}, CNN-Boosting~\cite{walach2016learning}, CP-CNN~\cite{sindagi2017generating}, CMTL~\cite{sindagi2017cnn}, Switching CNN~\cite{babu2017switching}, CSRNet~\cite{li2018csrnet}, CANNet~\cite{liu2019context}, SA-Net~\cite{cao2018scale}, ADSCNet~\cite{bai2020adaptive}, SASNet~\cite{song2021choose}, MLR~\cite{wu2006crowd}, KRR~\cite{an2007face}, Faster-OHEM-KCF~\cite{li2016deep}, LC-FCN~\cite{laradji2018blobs}, PSDNN~\cite{liu2019point}, AM-CNN~\cite{shao2018crowdhuman},  LSC-CNN~\cite{sam2020locate}, P2PNet~\cite{song2021rethinking}, Crowd-SDNet~\cite{wang2021self}, DPDNet~\cite{lian2021locating}}
\label{fig2}
\end{figure*}

In this section, according to the historical timeline review of the milestone methods outlined in Fig. \ref{fig2}, we pay attention to the related work in three aspects: the methods based on regression, the methods based on density estimation and the methods based on object detection.

\subsection{Regression based methods}
\label{subsection:regression_methods}

For highly crowded scenes, some researchers~\cite{chan2009bayesian, idrees2013multi} use regression methods to learn the mapping from images to the number of people. For example, Idrees et al.~\cite{idrees2013multi}, used sparse SIFT features to learn the support vector regression to estimate the number of people on each image patch. During training, these methods are usually accomplished by back propagation regression, and most of the loss functions leveraged are MAE. They firstly extract the global features of the image, and then use local features, such as gradient histogram features, texture features and edge features, etc. Finally they use linear regression,  Gaussian process regression, neural network regression and other algorithms to learn the mapping from the underlying features to the number of people. Ryan et al. made a clear comparison of these methods and features~\cite{ryan2015evaluation}. In crowd counting by regression, the problems of occlusion and background clutter are alleviated, nevertheless, the spatial information of the crowd is ignored. This will affect the accuracy of the model in some complex local regions.

\subsection{Density estimation-based methods}

To avoid the above global quantitative regression, researchers began to obtain a density map by exploring the mapping from local features in a small patch to the people density in the local area. Lemptisky et al.~\cite{lempitsky2010learning} first proposed to transform the point annotation into the GT of the density map through Gaussian kernel, and solved the above problems by learning the linear mapping between the local region feature map and the density map GT. Because of the limited ability of linear mapping to describe complex relationships, it is difficult to achieve ideal results. Pham et al.~\cite{pham2015count} employed a nonlinear mapping and used forests with split node layers for learning. This method also added a priori of congestion degree (Crowdedness Prior) to improve the accuracy of prediction and effectively reduce the forest.

In recent years, due to the strong representation advantage of CNN, more and more studies on crowd counting deploy CNN for density estimation. The commonly used network architecture is multi-branch. Walach et al.~\cite{walach2016learning} developed a regressor with cascaded CNNs. Images in different scales were sent to train multiple networks, and the output of these networks were fused to generate the final density map. Because each CNN has its own receptive field, they are only interested in targets of a certain density. In order to better capture context information, CSRNet~\cite{li2018csrnet} extended the kernel of CNN to obtain a larger receptive field so as to improve the predicting performance. CANNet \cite{liu2019context} and PCC net~\cite{gao2019pcc} encode and integrate image features from context and perspective respectively, obtaining better crowd counting performance. For all the above mentioned methods, their learning framework is supervised by a static density map, which is very sensitive to the labeling deviation and large-scale variation. There are some works for above problem, such as adaptive dilated convolution and self-correction supervision~\cite{bai2020adaptive},  Scale-Adaptive Selection Network~\cite{song2021choose} and Learning to Scale module~\cite{xu2022autoscale}. To obtain better views, video streams and cross-view are introduced by \cite{liu2020estimating} and \cite{zhang2021cross}, which improves counting performance for occluded people or those in low resolution. In addition, Shu et al.~\cite{shu2022crowd} studied a new direction, the difference between ground truth and predicted density map can be better measured by converting density mapping to frequency domain.

\subsection{Detection based methods}

Most early crowd counting studies focused on detection based methods\cite{enzweiler2008monocular, lin2010shape}, which use a window sliding method to detect pedestrians and calculate the total number of people. Pedestrian detection begins with detecting the whole body\cite{dalal2005histograms}. The detector is a sliding window classifier learned from hand-crafted features such as HoG, Haar wavelets and so on. In sparse scenes, these methods achieve good results. However, they perform poorly in crowded and heavily obscured scenes, where parts of the human body are invisible and object features are often missing. In order to deal with this difficulty, researchers proposed part-based and shape-based detectors~\cite{wu2007detection}. For example, Subburaman et al. used gradient orientation features to locate the head region~\cite{subburaman2012counting}, while Zeng et al. combined HoG and LBP feature learning to obtain a head-shoulder detector ~\cite{zeng2010robust}.
These early methods work by using handcrafted features. When the scene changes greatly or the crowd is highly dense, the performance of them is limited because of the weak ability to generalize the overall or partial appearance of human. The deep neural network developed in recent years is an effective way to address this weakness, but the shortcomings such as sample distribution and limited annotation, as well as parameter learning will be new difficulties to be overcome.

From the chronology in Fig. \ref{fig2}, it is observed that in the past two years, researchers have begun to explore new crowd counting methods from the point of view of how to improve the accuracy of head detection. However, the current head detection scheme has not successfully challenged the SOTA methods based on density estimation. Therefore, this paper attempts to contribute to head detection based crowd counting.

At present, deep learning based object detection can be classified into two-stage detection and one-stage detection. In the two-stage detectors, the most representative one is Region-CNN (RCNN)~\cite{zhang2018occlusion}. At the first stage, a regional proposal network is used to generate a set of proposals containing the bounding boxes of all objects. The second stage consists of a convolutional neural network classifier, which extracts features from each candidate box to represent the target and expands the receptive field through the downsampling process can increase the neighborhood spatial information. In recent years, many variants of RCNN have been applied to crowd counting, such as Faster RCNN~\cite{ren2015faster} and Mask RCNN~\cite{he2017mask}. In addition to two-stage detection, the one-stage detectors are proposed to directly estimate the bounding box of targets without region proposal, which greatly improves the efficiency but decreases the accuracy. Since the sample imbalance is a key cause of the accuracy decrease, one of the solutions is to suppress the loss contributions from the samples well classified in training. For this purpose, Lin et al. modified the cross-entropy loss function~\cite{lin2017focal}, named Focal Loss, the efficiency and accuracy were both higher than the best two-stage detectors at that time.


Although the existing two-stage and one-stage anchor-based detectors have recently gained great improvement, they still have some obvious shortcomings: (1) the manual anchor boxes’ scale and aspect ratio are with poor adaptive ability; (2) high recall anchor proposal in the highly-dense case is in high demand for computing resources; and (3) most of the anchor boxes are negative samples, leading to a severe sample imbalance. To address these problems, a series of anchor-free methods are proposed~\cite{chen2023perception, zhou2019objects, wang2019region, law2018cornernet}. Among them, CenterNet~\cite{zhou2019objects} estimated the object center and regressed to other attributes of objects. FCOS~\cite{tian2019fcos} showed that a center-ness branch could reduce the weight of low-quality bounding boxes, so that the accuracy of FCN-based detectors is higher than that of anchor-based detectors. FoveaBox~\cite{kong2020foveabox} used a category-sensitive semantic map to directly predict the confidence of existence.

However, these methods can not be directly applied to crowd counting for two main reasons. First, current datasets are not supported to detection based methods,  because most of the existing public datasets only annotate the person's head location as point instead of bounding box which is mandatory for the current mainstream object detection networks. Second, the canonical object detectors can not well sense the changes of the texture features in the highly dense head region because the sample imbalance is not well addressed. This makes the pedestrian head detection in complex scenes have a high rate of missed detection.

\section{Mask Focal Loss}
\label{section:mask_focal_loss}
In this section, we derive our proposed Mask Focal Loss from Focal Loss and its variants.

\subsection{Focal Loss and its variants}
\label{subsection:focal_variantes}

In the field of object detection, cross entropy is the most commonly used loss function. As discussed above, the prominent problem limiting the one-stage detectors in dense target detection is the sample imbalance. To solve this, Lin et al. proposed Focal Loss, a variant of cross entropy that can dynamically adjust the sample weight during training~\cite{lin2017focal}. Consider the binary classification, and set $c \in \left\{0,1\right\}$ as the class label: $0$ represents background and $1$ denotes object. Besides, $p$ is the probability of the predicted label as object, Focal Loss is defined as,

\begin{equation}
\label{eqn1}
{FL} = -(1-p_t)^\gamma \log(p_t),
\end{equation}
where $p_t=p$, if $c=1$, and $p_t=1-p$, if $c=0$; $\gamma>0$ is a controllable parameter. They defined $(1-p_t)^\gamma$  as a penalty factor for dynamically modulating the sample weights in the loss. For the well-classified samples, the penalty factor is close to 0, greatly reducing the sample weight for the loss. For the misclassified samples, the penalty factor is close to $1$, which is consistent with the original cross entropy. By adjusting the weights of samples for the loss, Focal Loss can greatly weaken the loss contributions of well-classified negative samples so as to alleviate the imbalance between positive and negative samples. Further, many excellent detection networks, e.g., YOLO, FCOS~\cite{tian2019fcos}, FoveaBox~\cite{kong2020foveabox}, generated feature maps based on the regions of anchor box, and used a Focal Loss variant, namely $\alpha$-Focal Loss~\cite{lin2017focal}, 

\begin{equation}
\label{eqn2}
L_{{f}}=-\frac{\alpha}{N} \sum_{x,y} \begin{cases}\left(1-\hat{p}_{x y}\right)^{\gamma} \log \left(\hat{p}_{x y}\right) & p_{x y}=1, \\ \left(\hat{p}_{x y}\right)^{\gamma} \log \left(1-\hat{p}_{x y}\right) & p_{x y}=0,\end{cases}
\end{equation}
where $x$ and $y$ represent the pixel coordinates in the image. $\hat{p}_{xy}$  is the predicted value. The confidence of being target at $(x,y)$ is $1$ ($p_{xy}=1$), and the confidence of being background at $(x,y)$ is $0$ ($p_{xy}=0$). $N$ represents the number of objects in the image, and $\alpha$ and  $\gamma$ are controllable hyper-parameters.

Subsequently, Zhou et al. proposed CenterNet for 3D object detection~\cite{zhou2019objects}. The main idea is to leverage key points instead of box anchors. The CenterNet method generates the heatmap of the target areas through the Gaussian kernel during training, and uses the pixel-wise logistic regression with Focal Loss. This variant of Focal Loss can be defined as,

\begin{equation}
\label{eqn3}
L_{{h}}=-\frac{\alpha}{N} \sum_{x,y} \begin{cases}\left(1-\hat{p}_{x y}\right)^{\gamma} \log \left(\hat{p}_{x y}\right) & p_{x y}=1, \\ \left(1-p_{xy}\right)^\beta\left(\hat{p}_{x y}\right)^{\gamma} \log \left(1-\hat{p}_{x y}\right) & p_{x y} \in [0,1).\end{cases}
\end{equation}

The point where $p_{xy}=1$ is the center of Gaussian kernel at $(x, y)$. It is also the center point of the pedestrian’s head in our application. $p_{xy}\in (0,1)$ is the heatmap value of the neighborhood of the key point, depending on the distance from the key point. The region where $p_{xy}=0$ represents the background; $\hat{p}_{xy}$ is the predicted heatmap value at $(x, y)$; $\alpha$, $\beta$ and $\gamma$ are predefined parameters. What is different from the original Focal Loss is that in the neighborhood of the key point, it considers a weight factor $(1-p_{xy})^\beta$ to further adjust the loss contributions of samples \cite{law2018cornernet}: closer to the center, larger $p_{xy}$ and less contributions.

Recently, Leng et al. proposed Poly loss, which approximates the cross entropy and Focal Loss via Taylor expansion~\cite{leng2022polyloss}. For example, to reduce the term number used for approximation, a coefficient perturbation method is adopted. Then, Focal Loss can be represented by the following poly-1 form,
\begin{equation}
\label{eqn4}
\begin{aligned}
FL^{P1}&=\left(1+\varepsilon_{1}\right)\left(1-p_{t}\right)^{1+\gamma}+1 / 2\left(1-p_{t}\right)^{2+\gamma}+\ldots \\ &=-\left(1-p_{t}\right)^{\gamma} \log \left(p_{t}\right)+\varepsilon_{1}\left(1-p_{t}\right)^{1+\gamma}+\ldots,
\end{aligned}
\end{equation}
where $p_t$ indicates the predicted value, $\varepsilon_1$ is the perturbation coefficient of ($1-p_t$) and $\gamma$ is predetermined parameters in original focal loss. A large number of experiments have proved the effectiveness of poly-1 form, that is, only introducing the first order polynomial of Focal Loss via Taylor expansion with some hyper-parameters can obtain significant accuracy improvement. 

Focal loss based on pixel-wise logistic regression under the poly loss framework can be modified as

\begin{equation}
\label{eqn5}
L_{{f}}^{P1}=-\frac{\alpha}{N} \sum_{x,y} \begin{cases}\left(1-\hat{p}_{x y}\right)^{\gamma} \log \left(\hat{p}_{x y}\right)-\left(1-\hat{p}_{x y}\right)^{\gamma+1} & p_{x y}=1, \\ \left(1-p_{x y}\right)^{\beta}\left(\left(\hat{p}_{x y}\right)^{\gamma} \log \left(1-\hat{p}_{x y}\right)-\left(\hat{p}_{x y}\right)^{\gamma+1}\right) & p_{x y} \in[0,1).\end{cases}
\end{equation}

\subsection{Mask Focal Loss}

There have been some existing popular work on improving the focal loss. \cite{li2020generalized} introduces the Generalized Focal Loss (GFocal) to overcome challenges in dense object detection, including inconsistent usage of quality estimation and classification, inflexible representation of bounding boxes, and lack of explicit guidance, resulting in improved representation learning for efficient and accurate detection. Besides, \cite{yeung2022unified} proposes the Unified Focal loss, a novel hierarchical framework that generalizes existing loss functions to effectively handle class imbalance in medical image segmentation. However, they do not address specific issues in the area of head detection. Compared to them, our mask focal loss focuses on solving the specific problem of the large-scale variation and detection confusion in dense crowd scenarios. We try to introduce the Gaussian heatmap generated from the head center points and their bounding box. For this kind of heatmap, it is a challenge to define a proper loss function for crowd counting network. If we use traditional $\alpha$-focal loss (Eq.(\ref{eqn2})) which takes all pixels in the bounding box as the positive samples, the head neighborhood area in the rectangular boundary box will have the same contribution as the center pixels. In this way, the loss contributions of the head neighborhood area is overemphasized, which will affect the model's accurate perception of the pedestrians' heads. If Eq. (\ref{eqn3}) is adopted, the pixels in the neighborhood of the head center will be taken as negative samples, which ignores the spatial coherence in the heatmap. In order to obtain a better loss function, we should consider the following two issues:

\textit{A. How to define positive and negative sample points properly?}

\textit{B. How to define the loss contributions of different kinds of samples, and dynamically adjust them?}

\begin{figure*}
\centering
\includegraphics[width=5.2in]{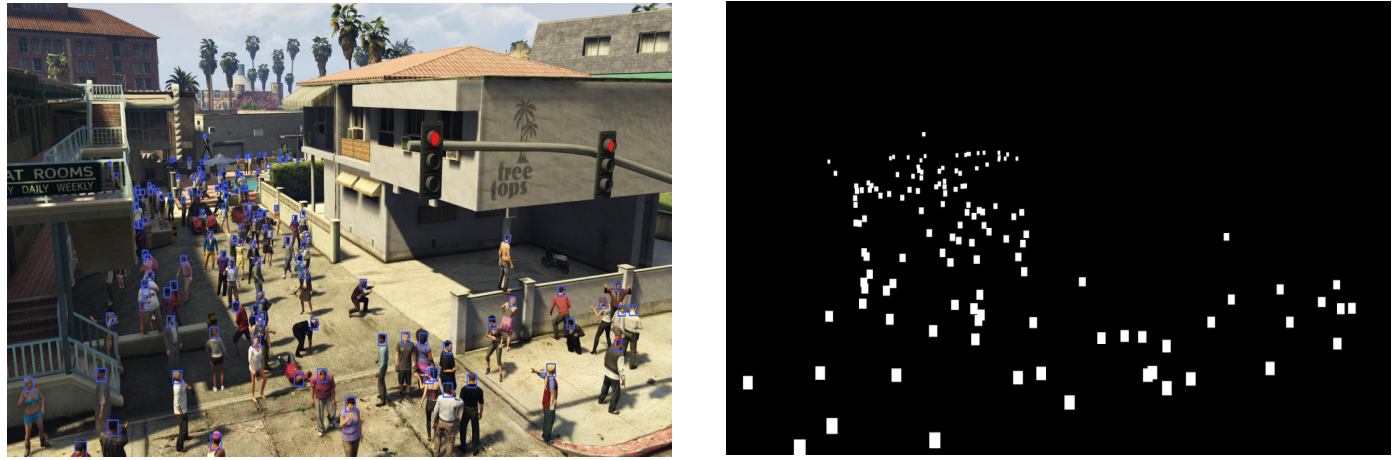}
\caption{For Mask Focal Loss, a target based mask map (right) can be generated from the annotations (left) during training. The mask value in the target bounding box is 1, and that in the background area is 0.}
\label{fig3}
\end{figure*}

\begin{figure}[!t]
\centering
\includegraphics[trim=0 190 390 0, width=4in]{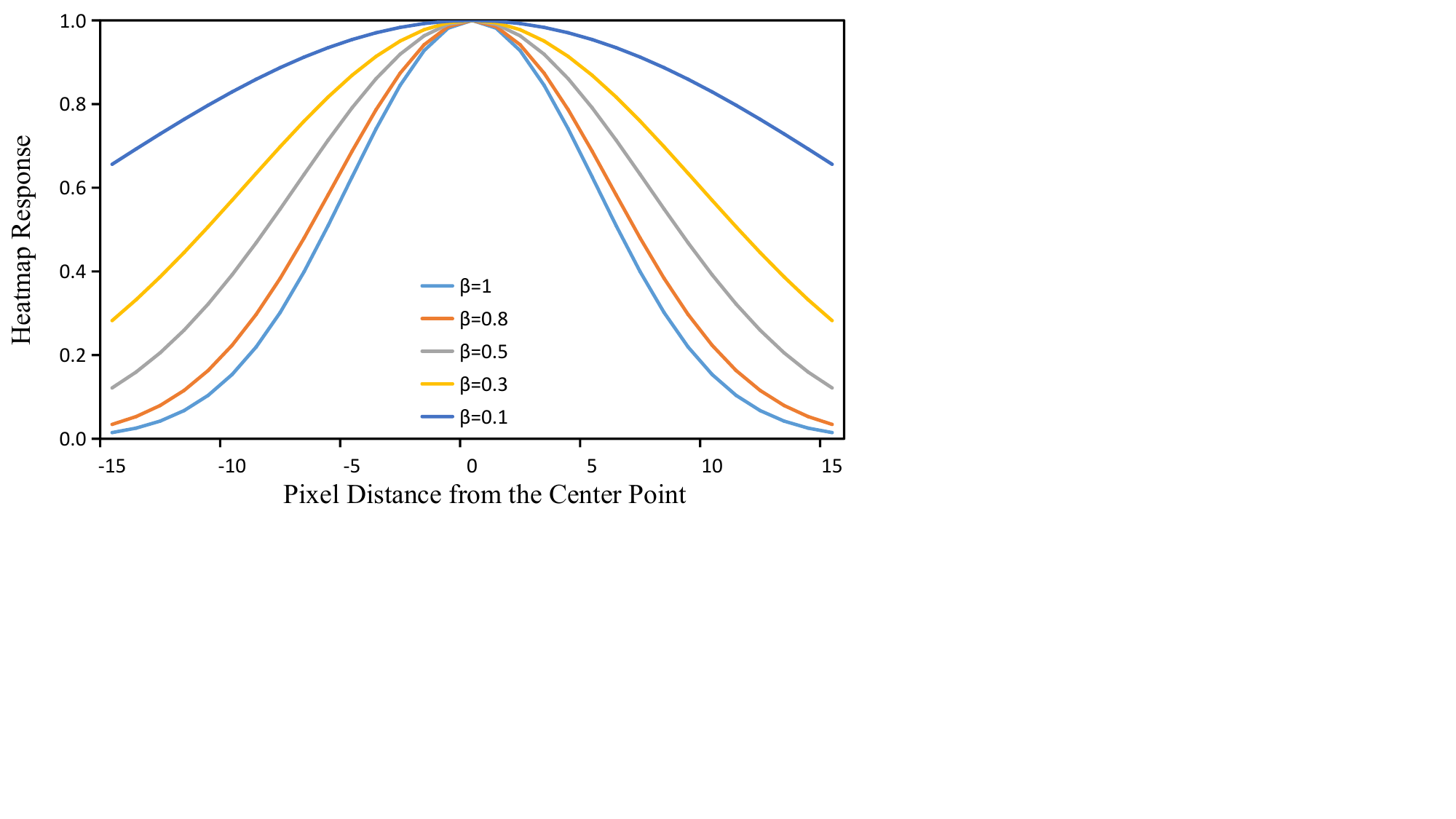}
\caption{Different $\beta$ leads to different contributions to the total loss.}
\label{fig4}
\end{figure}

For Issue A, spatial coherence is very important for the model perception of heads, especially small heads in crowded scenes. Therefore, for the pixels in the neighborhood of head centers, their expectation should be larger than 0 when calculating heatmap loss. We introduce an area mask (see in Fig. \ref{fig3}), and set the whole area inside the head bounding boxes as positive samples (mask=1) and the outside one as negative samples (mask=0). The division of positive and negative samples is convenient for us to treat the loss function differently.

For Issue B, we have known that the exceptions $\in (0,1)$ for the positive samples outside the head centers.
As distinguished from the loss used in CenterNet (Eq. (\ref{eqn3})), we use $\Delta p = \left|p_{x y}-\hat{p}_{x y}\right|$ instead of $1-\hat{p}_{xy}$. $\Delta p$ accurately illustrates the prediction error from the heatmap. In addition, we use $(p_{xy})^\beta$ to adaptively adjust the loss contributions as shown in Fig. \ref{fig4}. The loss of negative samples ($\text{mask} = 0$) is the same as the original Focal Loss (Eq. (\ref{eqn2})) since their exceptions are all 0. Consequently, we formulate the mask focal loss as
\begin{equation}
\small{
\label{eqn6}
L_{{m}}=-\frac{\alpha}{N} \sum_{x,y} \begin{cases}\left(p_{x y}\right)^{\beta} \Delta {p}^{\gamma} \log (1-\Delta {p}),  & \text {mask}=1, \\
\left(\hat{p}_{x y}\right)^{\gamma} \log \left(1-\hat{p}_{x y}\right), & \text {mask}=0. 
\end{cases}
}
\end{equation}
where $\beta$ and $\gamma$ is predefined parameters. In fact, $\text{mask}=1$ lies in where $p_{xy}\in (0,1]$ and $\text{mask}=0$ exactly when $p_{xy}=0$.
The absolute value $\left|p_{x y}-\hat{p}_{x y}\right|$ is used to ensure non-negativity and make the value within the range of $[0,1]$. It is obvious that in the positive sample region ($\text { mask }=1, p_{xy} \in (0,1]$), the greater the difference between the prediction and the heatmap the greater $\Delta p^\gamma$, meaning more contributions to the total loss.

To facilitate comparison, the characteristics of $L_f$, $L_h$ and $L_m$ are shown in Table \ref{table2}. It is obvious that, on the one hand, when $\beta = 0$ and the heatmap ground truth is binary feature map-based, $L_h$ and $L_m$ both reduce to $L_f$. On the other hand, when $\text{mask = 0}$, $p_{xy}=0$, then the second term $(1-p_{xy})^\beta(\hat{p}_{xy})^\gamma\log(1-\hat{p}_{xy})$ in $L_m$ is same as the second term in $L_f$. With this observation, when the mask used by $L_m$ is set to the key point, $L_m$ completely reduces to $L_h$.

\begin{table}
\centering
\caption{Comparison on three loss functions}
\label{table2}
\renewcommand\arraystretch{2}

\begin{tabular}{cccc}
\hline
\multicolumn{1}{c}{Type}      &  \ Featuremap based &    Heatmap based    & Mask (ours)      \\ \hline
\vspace{-12pt} &&& \\ 
\vspace{6pt} \begin{tabular}[c]{@{}l@{}}Heatmap/Binary \\ Feature Map\end{tabular}  
& \adjustbox{valign=c}{\includegraphics[width=2cm]{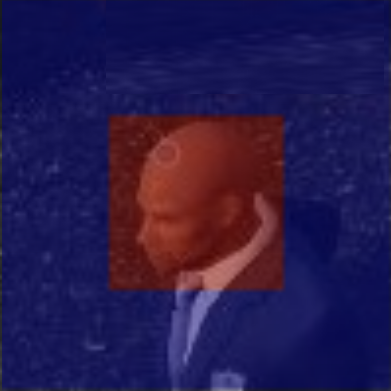}} 
& \adjustbox{valign=c}{\includegraphics[width=2cm]{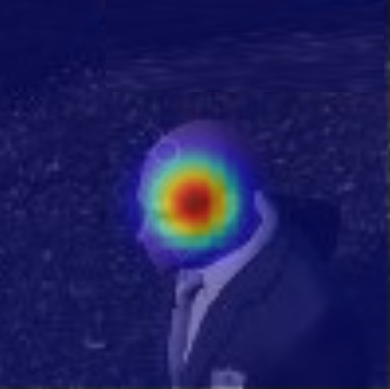}} 
& \adjustbox{valign=c}{\includegraphics[width=2cm]{heatfeatmask1.png}} \\
\hline
\vspace{-12pt} &&& \\ 
\vspace{6pt} Mask 
& \adjustbox{valign=m}{\includegraphics[width=2cm]{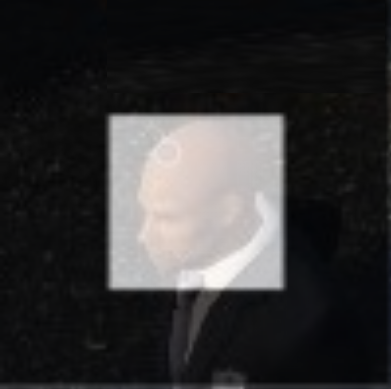}} 
& \adjustbox{valign=m}{\includegraphics[width=2cm]{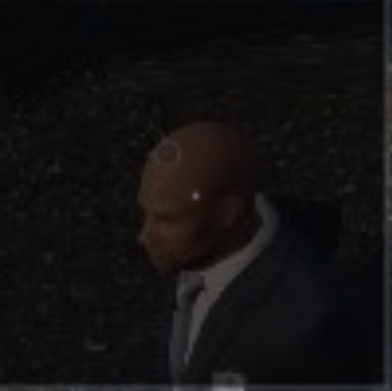}} 
& \adjustbox{valign=m}{\includegraphics[width=2cm]{heatfeatmask2.png}} \\ \hline

Loss Function & $L_{{f}}$ & $L_{{h}}$ & $L_{{m}}$ \\ \hline
Applicable to & \begin{tabular}[c]{@{}l@{}}FCOS, \\ RetinaNet,\\ FoveaBox,\\ YOLO\end{tabular} 
& \begin{tabular}[c]{@{}l@{}}CenterNet,\\ CornerNet\end{tabular} 
& \begin{tabular}[c]{@{}l@{}}All \\ detectors\end{tabular} \\
\hline
\end{tabular}
\end{table}

Therefore, the mask focal loss is also a kind of unified Focal Loss, which unifies the loss calculation methods based on heatmap and binary feature map. The loss based on binary feature map is applicable to detectors such as FCOS, RetinaNet, FoveaBox, YOLO, and the loss based on heatmap is only limited to CenterNet and CornerNet. On the contrary, our proposed Mask Focal Loss is applicable to all of these canonical detectors, making them easily applied to various datasets of crowd counting and providing a strong basis for surpassing the crowd counting methods based on density estimation.

As discussed in Subsection \ref{subsection:focal_variantes}, a Mask Focal Loss based on the poly loss framework (Eq. (\ref{eqn4})) can be yielded as
\begin{equation}
L_{{m}}^{P1}=-\frac{\alpha}{N} \sum_{x,y} 
\begin{cases}{\Delta p}^{\gamma} \log (1-\Delta p)- (p_{xy})^\beta {\Delta p}^{\gamma+1}, & \text{mask}=1 \quad (\text{where}\  p_{x y} \in (0,1]), \\ (\hat{p}_{xy})^{\gamma} \log \left(1-\hat{p}_{xy}\right)-\left(\hat{p}_{x y}\right)^{\gamma+1}, & \text{mask}=0\quad (\text{where}\  p_{xy}=0). \end{cases}
\label{eqn7}
\end{equation}
In the following experiments, we use this variant of Mask Focal Loss for comparison.

\section{Experiments and results}
\label{section:experiments_results}

In this section, we evaluate the proposed loss function. Firstly, we public a synthetic dataset, called GTA\_Head, with bounding box annotations suitable for the canonical object detection networks. Secondly, we use the mask focal loss with CenterNet for obtaining the optimal parameters $\beta$ and $\gamma$ through an ablation study on the GTA\_Head dataset. After that, the experimental results are compared with other canonical models on GTA\_Head, SCUT-Head~\cite{peng2018detecting} and CroHD~\cite{sundararaman2021tracking} datasets.

\subsection{GTA\_Head dataset}
\label{subsection:GTA_head_dataset}

In order to better train and evaluate dense crowd counting networks, we build a dataset for head detection. Following~\cite{wang2019learning}, the virtual world game Grand Theft Auto V (GTA5) is used to create the synthetic dataset due to the Rockstar Advanced Game Engine (RAGE) with excellent performance at the scene rendering, character modeling, texture details and weather effects and with the advantages of high image quality, easy labeling and scene controllability. The existing synthetic datasets limit the application of the detection models because they do not meet the requirements of the bounding box annotation. Therefore, we build in this subsection a synthetic dataset GTA\_Head, and provide a fast semi-automatic labeling method. The dataset is publicly available at \url{https://github.com/gkw0010/GTAV_Head-dataset}.

\begin{figure*}[!t]
\centering
\includegraphics[width=5.2in]{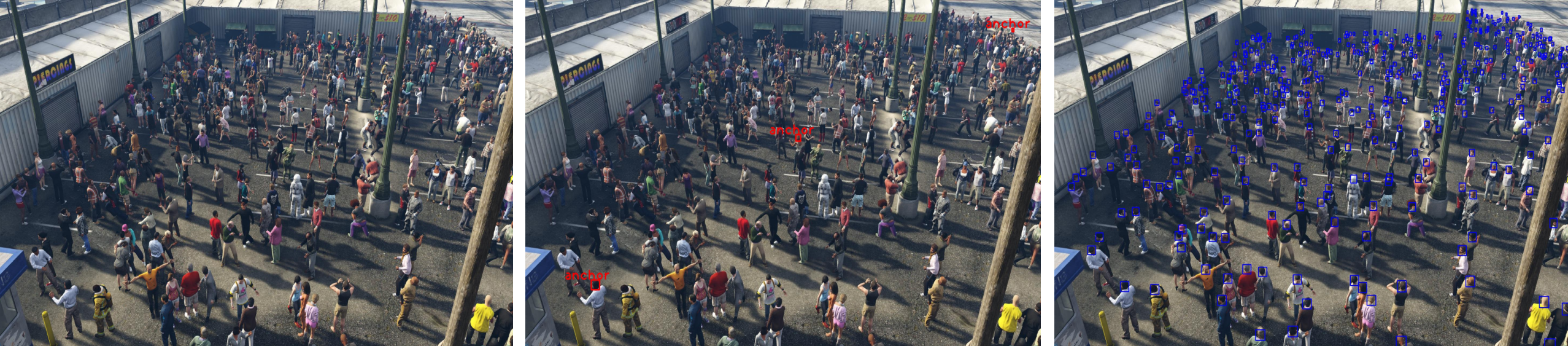}
\caption{Given a scene (left), with manually setting several representative anchor boxes (mid) , all pedestrian head sizes can be obtained through linear transformation in an automatic way (right).}
\label{fig5}
\end{figure*}

In our synthetic dataset GTA\_Head, the target center coordinates are taken from GTA5 through GCC dataset collector and labeler (GCC-CL) program based on Script Hook V script~\cite{wang2019learning}. In order to obtain the bounding boxes of the pedestrian head, we design a semi-automatic anchor-based annotation method. According to the head size change (as shown in Fig. \ref{fig5}), we manually preset bounding boxes for $N_b$ typical heads. The size of a head box is determined by the linear interpolation of the distances between the head and its two closest preset boxes. Considering that the change of the pedestrian head size is consistent in the images of the same scene, that is, the pedestrian head sizes at the same position are mutually close in the images of the same scene. Therefore, one scene only needs a few bounding boxes to complete the generation of head annotation. 

\begin{figure}
\centering
\includegraphics[trim=-45 180 690 0, width=5.2in]{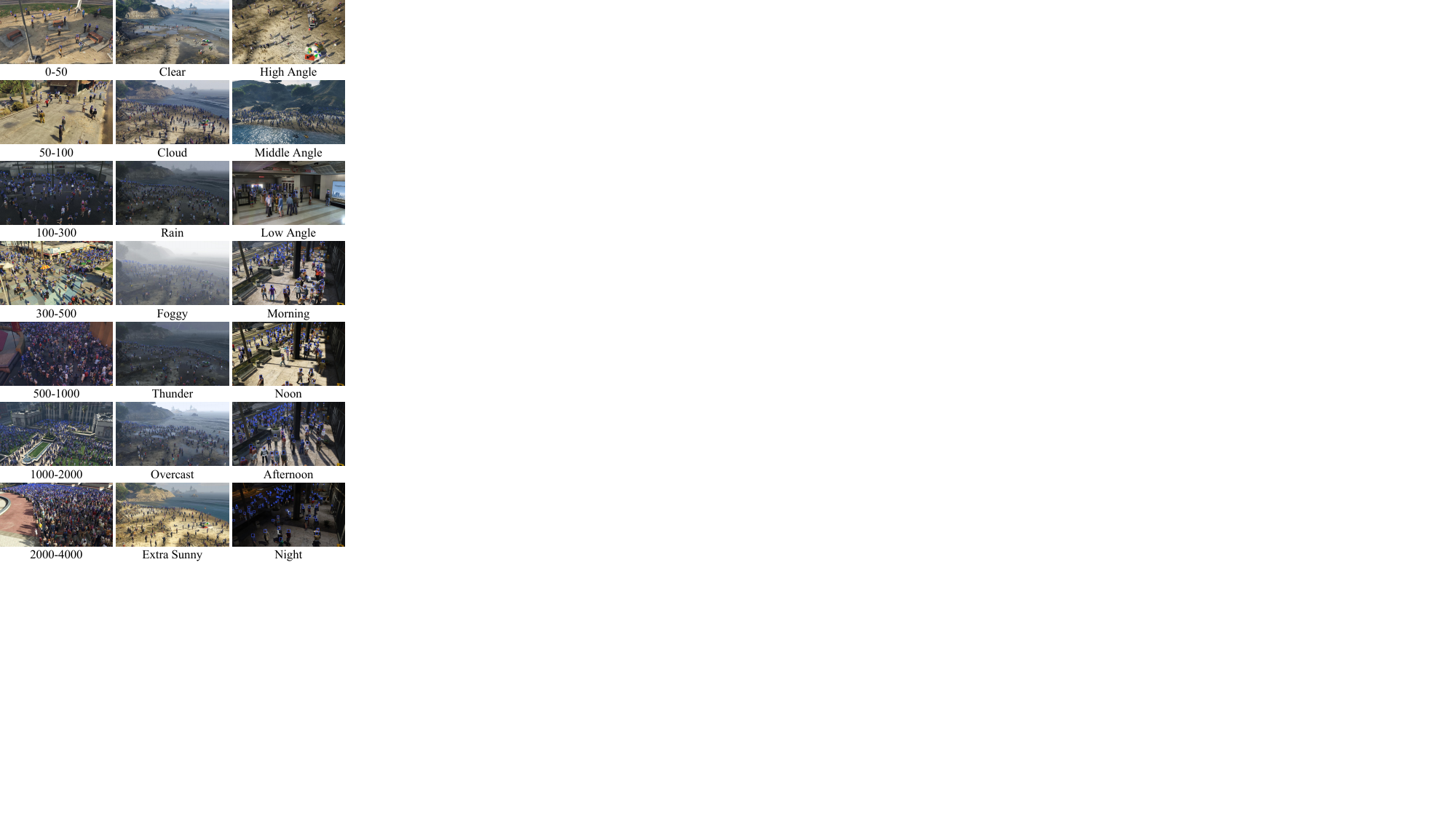}
    \caption{GTA\_Head dataset examples with different density(left), weather(mid), shooting angle and day part (right).}
    \label{fig6}
\end{figure}

Eventually, GTA\_Head dataset consists of 5096 images with 1732043 head bounding boxes. By contrast with other datasets, GTA\_Head provides pedestrian head annotations for a large number of complex scenes, including indoor shopping malls, subways, stadiums and squares, etc. To make the dataset maximally close to the real world and ensure the richness of the dataset for training and evaluation, we consider the cases with a variety of crowd densities, in different day parts, in different weathers, e.g., sunny, cloud, rain, fog, thunder, cloudy and exceptionally sunny, as well as at different shooting angles. Some samples are shown in Fig. \ref{fig6}.

\subsection{Evaluation metrics}

Like the methods based on density estimation, our evaluation metrics of crowd counting are MAE (Mean Absolute Error) and RMSE (Root Mean Squared Error). The following experiments will use these metrics defined as follows,

\begin{equation}
\label{eqn8}
MAE =\frac{1}{M} \sum_{i=1}^{M}\left|C_{i}-C_{i}^{G T}\right|,
\end{equation}

\begin{equation}
\label{eqn9}
{RMSE} =\sqrt{\frac{1}{M} \sum_{i=1}^{M}\left|C_{i}-C_{i}^{G T}\right|^{2}},
\end{equation}
where $M$ is the number of images in the test set, and $i$ indicates the $i$-th image sample considered; $C_i^{GT}$ is the ground truth of the number of people, and $C_i$ denotes the predicted quantity of the $i$-th image sample.

\subsection{Ablation study} 
\label{Ablation study}
\subsubsection{Configuration}

\begin{figure}[!t]
\centering
\includegraphics[trim=0 270 520 0, width=4.5in]{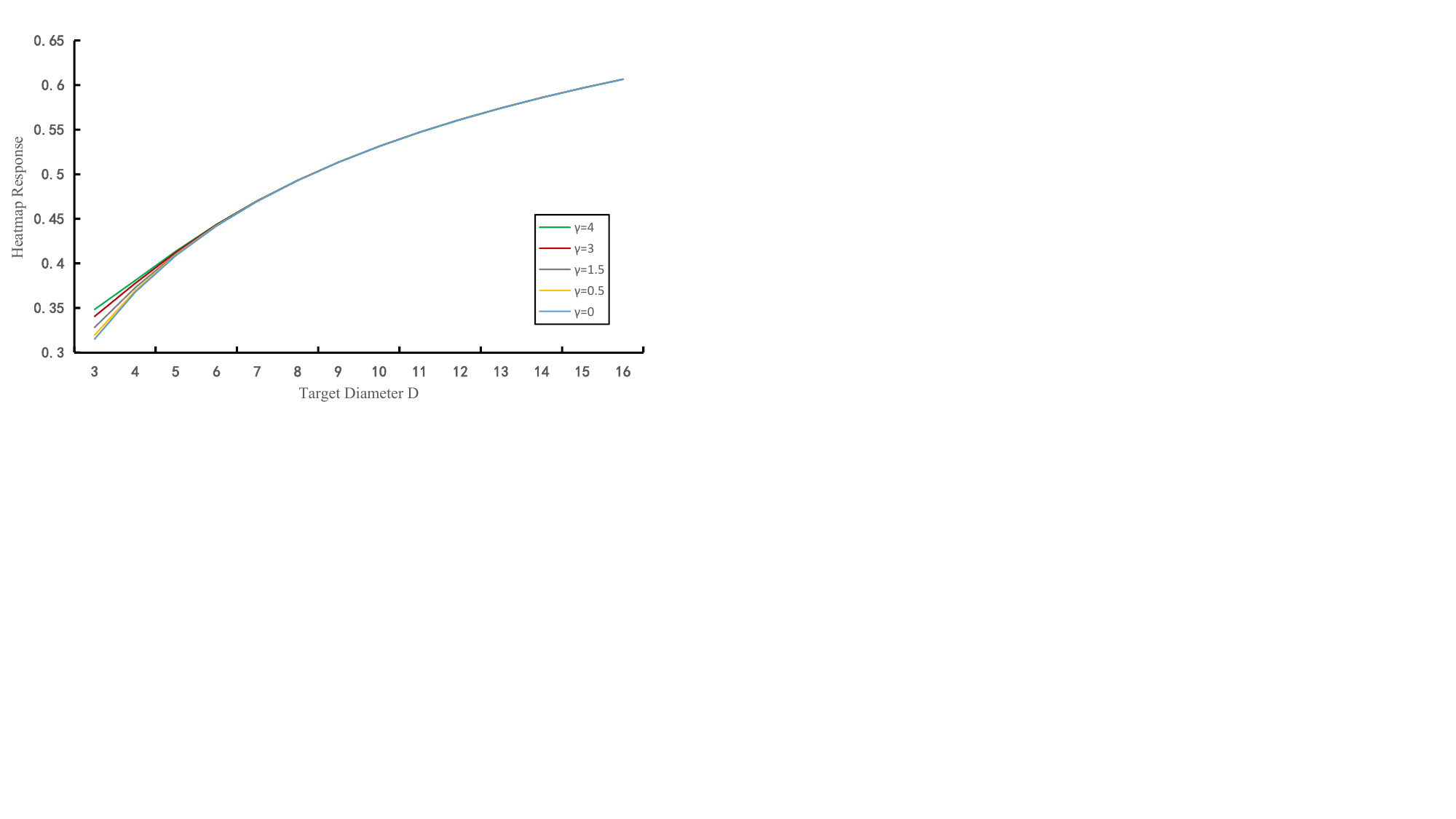}
\caption{The effect of small object sensing factor. Obviously, in the low value (small object) region of $D$, the heatmap value can be effectively enhanced through the modification of $\eta$ without affecting the heatmap response of large objects.}
\label{fig:sensing_factor}
\end{figure}

In order to find the optimal parameters of Mask Focal Loss, we performed an ablation experiment in this subsection. Since $\Delta p$  in Mask Focal Loss denotes the pixel-wise heatmap error, it can give full play to its potential in the field of heatmap classification. Based on this, the detection model CenterNet based on heatmap classification is a good choice. Compared with inefficient and wasteful calculation of potential objects enumerated by a large number of anchors, CenterNet creatively uses the anchor-free architecture and realizes end-to-end detection by predicting the center point and bounding box of objects, making detection more accurate, faster and simpler.


When generating the heatmap GT, for each box $b^i = (c_x^i, c_y^i, w^i, h^i)$, the heat value calculation function at point $(x, y)$ is $p_{xy}=\sum_i \exp ^{-{[(x-c_x^i)^2+(y-c_y^i)^2]}/{2\sigma^2}}$. 
The performance of the target detector relies heavily on the quality of the heatmap ground truth provided during training. For small targets that are difficult to detect, we need to give them a stronger heatmap response, enabling small target to have a larger loss contribution during model training.
Therefore, we modified $\sigma$ as below,

\begin{equation}
\label{eqn10}
{\sigma} =\dfrac{D\times (1+\eta\times \exp ^{-D})}{\varepsilon},
\end{equation}
where $D = 2\times\min(w, h)+1$ is radius of the target and $\varepsilon$ is a superparameter which is set to 6. $D\times (1+\eta\times \exp ^{-D})$ is the small target sensing factor, which can sense the radius of the target and delay the decay of the heatmap response of small targets from the center to the perimeter. Fig. \ref{fig:sensing_factor} shows the heatmap response at a fixed distance from the center point of different size targets.

In this experiment, the size of the input image is $1088 \times 608$. We randomly initialized the network at the beginning of training. The optimization algorithm is Adam, and the learning rate gradually decayed from $1e-2$ to $1e-4$; the batch size is set to $8$, and the number of training epochs is $36$. All experiments are carried out on a workstation with an Intel Xeon 16-core CPU (3.5GHz), 64GB RAM, and two GeForce RTX 3090Ti GPUs.

\subsubsection{Results and discussion}

Firstly, the CenterNet model with different loss functions and parameters is used to compare $L_h$ and $L_m$. For $L_h$, we set $\beta= 4$, $\gamma= 2$ as referred in~\cite{wang2019learning}.  For $L_m$, we set $\gamma \in \{1, 2, 3, 4, 5, 6\}$ for comparison.  Since the better MAE and RMSE are concentrated in $[3, 5]$, as shown in Table \ref{table3}, and the best MAE and RMSE are achieved at $\gamma=4$, we only show the results with $\gamma= 4$. It is observed that compared with $L_h$, the best performance of $L_m$ was achieved at $\gamma=4$, $\beta=0.25$, and the MAE was $36.99$.

\begin{table}[htbp]
\centering
\renewcommand\arraystretch{1.5}
\caption{The performance of $L_h$ and $L_m$ on GTA\_Head dataset ($\alpha=1$).}
\label{table3}
\begin{tabular}{>{\centering\arraybackslash}m{3cm} >{\centering\arraybackslash}m{3cm} >{\centering\arraybackslash}m{3cm}}
\hline
\textit{\textbf{Loss function}} & \textit{\textbf{MAE}} & \textit{\textbf{RMSE}} \\
\hline
$L_h(\gamma=2,\beta=4)${~\cite{zhou2019objects}} & $39.42$ & $90.80$ \\
$L_m(\gamma=4,\beta=0.25)$ & $\mathbf{36.99}$ & $88.34$ \\
$L_m(\gamma=4,\beta=0.375)$ & $37.66$ & $86.47$ \\
$L_m(\gamma=4,\beta=0.5)$ & $37.25$ & $\mathbf{85.84}$ \\
$L_m(\gamma=4,\beta=1.0)$ & $38.66$ & $89.73$ \\
\hline
\end{tabular}
\end{table}

Then, for the ablation experiment of $L^{P1}_m$ on CenterNet, we also set $\gamma= 1, 2, ..., 6$ for comparison. The experimental results of different $\gamma$ are shown in Table \ref{tab_ablation_a}. To control other variables, we set $\beta=0$. We can see that the appropriate increase $\gamma$ can improve detection performance. The best MAE ($36.15$) and RMSE ($84.75$) are obtained when $\gamma = 5$ and $6$, respectively. With larger $\gamma$, the loss contributions of the positive and negative samples can be adjusted more. Continuing to increase $\gamma$, the negative sample loss contributions will be too small, which is a state of reverse imbalance of positive and negative samples. In Table \ref{tab_ablation_b}, we set $\gamma$ according to the test results in Table \ref{tab_ablation_a}. Referring to Focal Loss~\cite{wang2019learning}, we set $\beta \in (0,1]$. 
It was observed that compared with $L_h$ (Table \ref{table3}), the mask focal loss ($L_m^{P1}$) using poly-1 form achieved the best performance when $\gamma= 4$, $\beta= 0.5$, where MAE and RMSE were $35.83$ and $83.75$, respectively. Because of the great improvement in the detection accuracy, we will use this set of parameters in the subsequent experiments.

\begin{table}
    \caption{Ablation study of $L_m^{P1}$ with the CenterNet model for different $\gamma  (\beta=0)$.}
    \label{tab_ablation_a}
    \centering
    \renewcommand\arraystretch{1.5}
    \begin{tabular}{>{\centering\arraybackslash}m{3cm} >{\centering\arraybackslash}m{3cm} >{\centering\arraybackslash}m{3cm}}
    \hline
    $\gamma$ & \textit{\textbf{MAE}} & \textit{\textbf{RMSE}} \\ \hline
    $1.0$                      & $45.72$               & $125.01$               \\
    $2.0$                      & $38.43$               & $92.47$                \\
    $3.0$                      & $37.65$               & $94.70$                \\
    $4.0$                      & $36.65$               & $88.87$             \\
    $5.0$                      & $\textbf{36.15}$               & $85.14$                \\
    $6.0$                      & $36.67$               & $\textbf{84.75}$                \\ \hline
    \end{tabular}

\end{table}
\begin{table}
\centering
\caption{Ablation study of $L_m^{P1}$ with the CenterNet model for different $\gamma$ and $\beta$.}
\label{tab_ablation_b}
        \centering
        \renewcommand\arraystretch{1.5}
    \begin{tabular}{>{\centering\arraybackslash}m{2.5cm} >{\centering\arraybackslash}m{2.5cm} >{\centering\arraybackslash}m{2.5cm} >{\centering\arraybackslash}m{2.5cm}}
    \hline
    $\beta$ & $\gamma$  & \textit{\textbf{MAE}} & \textit{\textbf{RMSE}} \\ \hline
    $0.3$       &   $3.0$        & $37.57$               & $89.95$               \\
    $0.5$       &   $3.0$        & $38.92$               & $93.62$                \\
    $0.3$       &   $4.0$        & $36.79$               & $87.84$                \\
    $0.5$       &   $4.0$        & $\textbf{35.83}$        & $\textbf{83.75}$     \\
    $0.75$      &   $4.0$        & $36.54$               & $88.74$                \\ 
    $1.0$       &   $4.0$        & $37.61$               & $88.32$                \\ 
    $0.3$       &   $5.0$        & $36.82$               & $86.24$                \\ 
    $0.5$       &   $5.0$        & $36.77$               & $87.08$                \\ 
    $1.0$       &   $5.0$        & $36.26$               & $85.21$                \\ 
    \hline
    \end{tabular}

\end{table}

\subsection{Evaluation and comparison on different networks}

\subsubsection{Implementation}

In order to evaluate the mask focal loss, we used three different datasets of the real world and virtual world for comparison: SCUT-Head, CroHD and our GTA\_Head, respectively. For the SCUT-Head,  it consists of two parts: Part\_A includes $2000$ images sampled in which $67321$ heads are marked; Part\_B has $2405$ images with $43930$ annotation boxes so that the experiment on this dataset will be divided into two parts.

For the selection of models, we investigate the density estimation-based models and the detection based models, respectively. We consider the SOTA networks, CSRNet and CANNet as the baselines of density estimation methods. For the detection based models, we focus on the one-stage detectors, and the performance of the two-stage detector (e.g., Faster R-CNN) is only for reference. For one-stage detectors, we choose two kinds of models based on heatmap and binary feature map: the network based on heatmap is CenterNet, which employs the loss functions of $L_h$ and $L_m^{P1}$, respectively; and the networks based on binary feature map include YOLOv5, RetinaNet, FCOS and FoveaBox, that use also $L_f$  and $L_m^{P1}$, respectively.

For all the models involved in the comparison, the parameters of Mask Focal Loss are set as $\gamma=4$, $\beta= 0.5$, which are obtained from ablation experiments. The configuration of the comparison experiment is the same as that of the ablation experiment. For the positive samples in the binary feature map, $p_{xy}=1$, $\beta$ will not have an impact on $(p_{xy})^\beta$ and will not be discussed.


\begin{figure*}[!t]
\centering
\includegraphics[width=5.2in, height=6.2in]{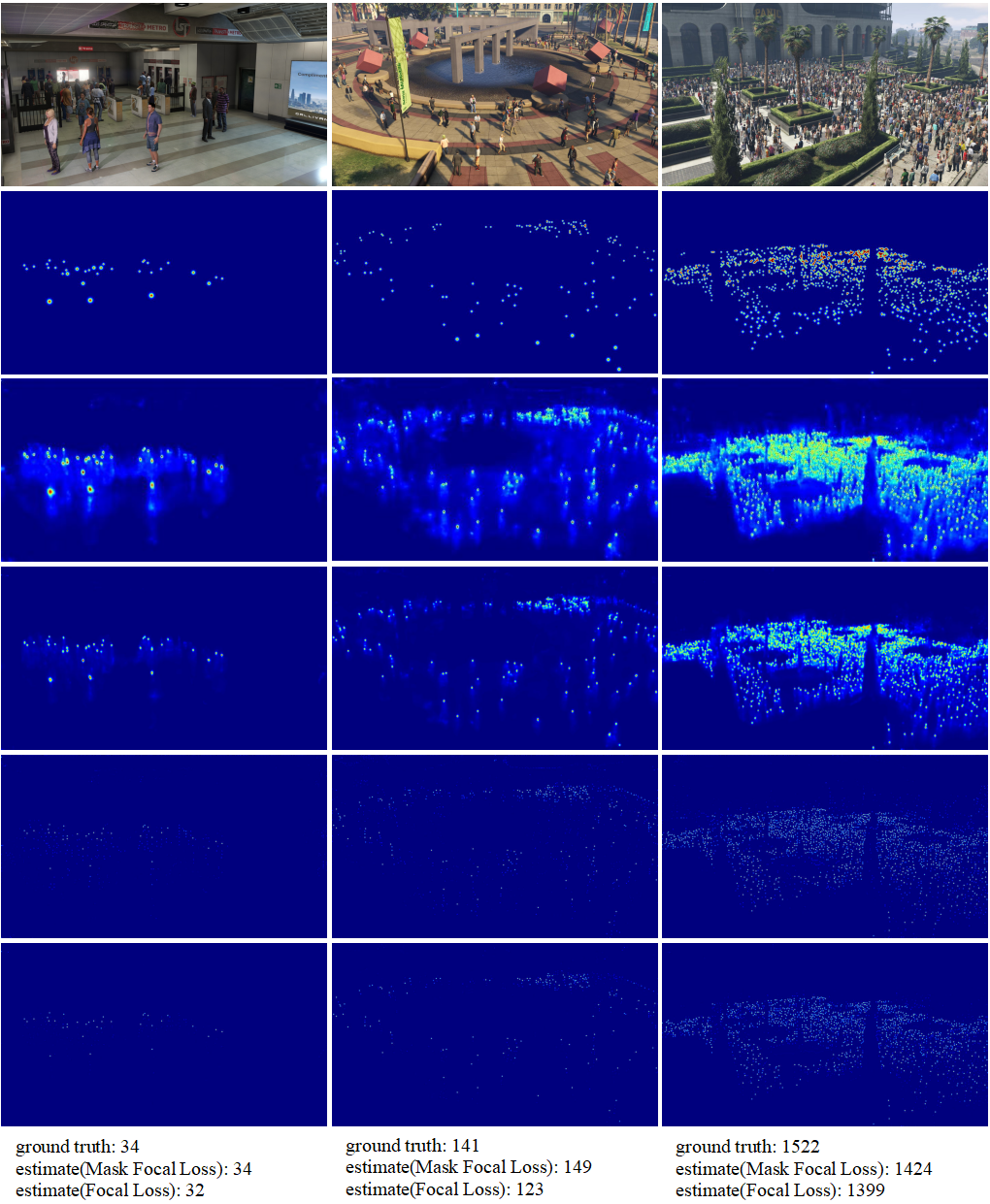}
\caption{Detection examples on GTA\_Head using CenterNet. The 1st row: original images; the 2nd row: heatmap ground truth of above original images; the 3rd and 4th rows: heatmaps generated with Mask Focal Loss and Focal Loss, respectively; the 5th and 6th rows: results of max pooling for heatmaps of the 3rd and 4th rows. In the process of analyzing the max-pooled heatmaps, specific points characterized by pixel values surpassing a predefined threshold are identified and marked as potential head locations. This threshold serves as a discriminative criterion, allowing the system to differentiate between significant heatmap activations and background noise.}
\label{fig7}
\end{figure*}

\begin{table*}[]
\hspace{-1cm}
\centering
\caption{Comparison of loss function on SCUT-Head. * indicates the model with the mask focal loss.}
\label{table6}
\renewcommand\arraystretch{1.5}
\begin{tabular}{c@{\hspace{0.15cm}}c@{\hspace{0.15cm}}c@{\hspace{0.15cm}}c@{\hspace{0.15cm}}c@{\hspace{0.15cm}}c@{\hspace{0.15cm}}c@{\hspace{0.15cm}}c}
\hline
\multirow{2}{*}{Model type}  & \multirow{2}{*}{}  & \multirow{2}{*}{Network} & \multirow{2}{*}{Backbone} & \multicolumn{2}{l}{SCUT-HEAD\_A}   & \multicolumn{2}{l}{SCUT-HEAD\_B}    \\ \cmidrule(lr){5-8}
   &       &        &        & MAE          & RMSE  & MAE           & RMSE                   \\ \hline
\multirow{11}{*}{\begin{tabular}[c]{@{}c@{}}Detection \\ based\end{tabular}} & \multirow{10}{*}{\begin{tabular}[c]{@{}c@{}}One \\ stage\end{tabular}} & CenterNet                & DLA-34                    & 1.92                   & 2.94                   & 2.23                   & 4.56                   \\
                                                                             &                                                                        & CenterNet*               & DLA-34                    & \textit{\textbf{1.26}} & \textit{\textbf{1.96}} & \textit{\textbf{1.78}} & \textit{\textbf{3.63}} \\ \cmidrule(lr){3-8} 
                                                                             &                                                                        & FCOS                     & ResNet50                  & 2.26                   & 3.38                   & 2.33                   & 5.23                   \\
                                                                             &                                                                        & FCOS*                    & ResNet50                  & \textbf{2.09}          & \textbf{3.05}          & \textbf{2.29}          & \textbf{5.19}          \\ \cmidrule(lr){3-8} 
                                                                             &                                                                        & FoveaBox                 & ResNet50                  & 1.92                   & 2.89                   & 2.52                   & 4.80                   \\
                                                                             &                                                                        & FoveaBox*                & ResNet50                  & \textbf{1.84}          & \textbf{2.80}          & \textbf{2.23}          & \textbf{4.61}          \\ \cmidrule(lr){3-8} 
                                                                             &                                                                        & RetinaNet                & ResNet50                  & 2.35                   & 3.40                   & 4.04                   & 8.45                   \\
                                                                             &                                                                        & RetinaNet*               & ResNet50                  & \textbf{1.90}          & \textbf{2.71}          & \textbf{2.13}          & \textbf{3.21}          \\ \cmidrule(lr){3-8} 
                                                                             &                                                                        & YOLOv5                   & YOLOv5x                   & 2.84                   & 4.04                   & 2.52                   & 5.20                   \\
                                                                             &                                                                        & YOLOv5*                  & YOLOv5x                   & \textbf{2.41}          & \textbf{3.53}          & \textbf{2.27}          & \textbf{4.97}          \\ \cmidrule(lr){2-8} 
                                                                             & \begin{tabular}[c]{@{}c@{}}Two \\ stage\end{tabular}                   & Faster-RCNN              & ResNet50                  & 2.89                   & 4.14                   & 2.96                   & 4.96                   \\ \hline
\multicolumn{2}{c}{\multirow{2}{*}{\begin{tabular}[c]{@{}c@{}}Density estimation \\ based\end{tabular}}}                                           & CSRNet                   & VGG-16                    & 2.46                   & 3.41                   & 3.75                   & 7.85                   \\ \cmidrule(lr){3-8} 
\multicolumn{2}{c}{}                                                                                                                                  & CANNet                   & VGG-16                    & 2.13                   & 2.81                   & 3.32                   & 6.26                   \\ \hline
\end{tabular}
\end{table*}

\begin{table}[]
\centering
\caption{Comparison of loss function on CroHD. * indicates the model with the mask focal loss.}
\label{table:CroHD_results}
\renewcommand\arraystretch{1.5}
\begin{tabular}{>{\centering\arraybackslash}m{1.5cm} >{\centering\arraybackslash}m{1.5cm} >{\centering\arraybackslash}m{1.5cm}>{\centering\arraybackslash}m{1.5cm}>{\centering\arraybackslash}m{1.5cm} >{\centering\arraybackslash}m{1.5cm}}
\hline
Model type                                                                   &                                                                        & Network     & Backbone & MAE                    & RMSE                   \\ \hline
\multirow{11}{*}{\begin{tabular}[c]{@{}c@{}}Detection \\ based\end{tabular}} & \multirow{10}{*}{\begin{tabular}[c]{@{}c@{}}One \\ stage\end{tabular}} & CenterNet   & DLA-34   & 8.11                   & 10.12                  \\
                                                                             &                                                                        & CenterNet*  & DLA-34   & \textit{\textbf{6.96}} & \textit{\textbf{8.72}} \\ \cmidrule(lr){3-6} 
                                                                             &                                                                        & FCOS        & ResNet50 & 6.87                   & 8.50                   \\
                                                                             &                                                                        & FCOS*       & ResNet50 & \textbf{6.32}          & \textbf{8.00}          \\ \cmidrule(lr){3-6} 
                                                                             &                                                                        & FoveaBox    & ResNet50 & 7.71                   & 9.77                   \\
                                                                             &                                                                        & FoveaBox*   & ResNet50 & \textbf{7.46}          & \textbf{9.40}          \\ \cmidrule(lr){3-6} 
                                                                             &                                                                        & RetinaNet   & ResNet50 & 7.13                   & 9.17                   \\
                                                                             &                                                                        & RetinaNet*  & ResNet50 & \textbf{6.26}          & \textbf{8.24}          \\ \cmidrule(lr){3-6} 
                                                                             &                                                                        & YOLOv5      & YOLOv5x  & 5.62                   & 7.10                   \\
                                                                             &                                                                        & YOLOv5*     & YOLOv5x  & \textbf{5.07}          & \textbf{6.40}          \\ \cmidrule(lr){2-6} 
                                                                             & \begin{tabular}[c]{@{}c@{}}Two \\ stage\end{tabular}                   & Faster-RCNN & ResNet50 & 8.88                   & 11.12                  \\ \hline
\multicolumn{2}{c}{\multirow{2}{*}{\begin{tabular}[c]{@{}c@{}}Density estimation \\ based\end{tabular}}}                                           & CSRNet      & VGG-16   & 5.25                   & 6.56                   \\ \cmidrule(lr){3-6} 
\multicolumn{2}{c}{}                                                                                                                                  & CANNet      & VGG-16   & 6.76                   & 8.12                   \\ \hline
\end{tabular}
\end{table}

\begin{table}[]
\centering
\caption{Comparison of loss function on GTA\_Head. * indicates the model with the mask focal loss.}
\label{table8}
\renewcommand\arraystretch{1.5}
\begin{tabular}{>{\centering\arraybackslash}m{2cm}>{\centering\arraybackslash}m{2cm}>{\centering\arraybackslash}m{2cm}>{\centering\arraybackslash}m{2cm} >{\centering\arraybackslash}m{2cm}}
\hline
Model type                                                                              & Network    & Backbone & MAE                     & RMSE                    \\ \hline
\multirow{10}{*}{\begin{tabular}[c]{@{}c@{}}Detection \\ based\end{tabular}}            & CenterNet  & DLA-34   & 39.42                   & 90.80                   \\
                                                                                        & CenterNet* & DLA-34   & \textbf{35.83}          & \textbf{83.75}          \\ \cmidrule(lr){2-5} 
                                                                                        & FCOS       & ResNet50 & 40.15                   & 79.11                   \\
                                                                                        & FCOS*      & ResNet50 & \textbf{38.92}          & \textbf{75.50}          \\ \cmidrule(lr){2-5} 
                                                                                        & FoveaBox   & ResNet50 & 34.69                   & 64.40                   \\
                                                                                        & FoveaBox*  & ResNet50 & \textbf{32.40} & \textbf{59.66} \\ \cmidrule(lr){2-5} 
                                                                                        & RetinaNet  & ResNet50 & 48.60                   & 106.58                  \\
                                                                                        & RetinaNet* & ResNet50 & \textbf{44.63}          & \textbf{92.61}          \\ \cmidrule(lr){2-5} 
                                                                                        & YOLOv5     & YOLOv5x  & 67.17                   & 177.58                  \\
                                                                                        & YOLOv5*    & YOLOv5x  & \textbf{64.20}          & \textbf{165.78}         \\ \hline
\multirow{2}{*}{\begin{tabular}[c]{@{}c@{}}Density estimation \\  based\end{tabular}} & CSRNet     & VGG-16   & 65.20                   & 92.47                   \\ \cmidrule(lr){2-5} 
                                                                                        & CANNet     & VGG-16   & 60.10                   & 80.36                   \\ \hline
\end{tabular}
\end{table}

\subsubsection{Results} 

The results of SCUT-Head are shown in Table \ref{table6}. We can see that the accuracy of counting has been improved for all networks by replacing focal loss with our mask focal loss. Compared to traditional two-stage detectors and density estimation methods, one-stage detectors have better performance in low-density and complexity scenarios. CenterNet exhibits the greatest improvement and counting performance, with its MAE and RMSE reduced by 34.38\% and 33.33\%, respectively, over SCUT-HEAD\_A. On SCUT-HEAD\_B, the improvement is 20.18\% and 20.39\%, respectively. Therefore, the mask focal loss is more effective for enhancing the heat map based detection model. In addition to this, the MAE and RMSE of the other one-stage models are improved by 1.72\%-47.03\% and 0.76\%-61.99\%, correspondingly.

In Table~\ref{table:CroHD_results}, the one-stage detection models consistently exhibit substantial improvements over their non-mask focal loss counterparts. Specifically, FCOS* and YOLOv5* stand out with a notable counting performance. YOLOv5* achieves a significant 9.16\% reduction in MAE and an 9.86\% reduction in RMSE, affirming the positive impact of the mask focal loss. Furthermore, the one-stage models generally outperform the two-stage Faster-RCNN, emphasizing the efficacy of these architectures on the CroHD dataset. For density estimation, both CSRNet and CANNet demonstrate competitive performance, which indicates their potential in high-density scenarios.

For GTA\_Head, our experiment did not consider any two-stage detector because the dataset with ultra-high density samples needs a large number of area proposals which require extremely large GPU resources. Fig. \ref{fig7} shows the generated heatmaps and counting results with different loss functions. We can find that the perception ability of head has been greatly improved after using the mask focal loss. Mask Focal Loss makes the valid head size in the generated heatmap larger. It is obvious in the scene with high crowd density, reducing the missed detection and improving the detecting accuracy. The count of identified points corresponds to the estimated number of heads within the given image or region of interest. This approach leverages the pooling operation to emphasize regions with maximal activation, aiding in the accurate localization and enumeration of heads based on the intensity of heatmap values. The test results are shown in Table \ref{table8}. With the mask focal loss, all of the counting accuracies of the one-stage detectors are improved. In addition, we can see that the performance of the density estimation-based methods on GTA\_Head is far behind one-stage detectors. FoveaBox* particularly stands out, showcasing a notable 32.40 in MAE and 59.66 in RMSE. Density estimation models face challenges in precisely localizing instances, as evidenced by their higher errors compared to detection-based models. This is sufficient to demonstrate the advantages of the detection based models in crowd counting in complex scenes.

In a word, Mask Focal Loss can at the same time apply to any kind of canonical detection network and effectively improve the performance of detectors in the field of crowd counting. With Mask Focal Loss, the detection based methods outperform the SOTA methods based on density estimation. In addition, we also find that anchor-free is more competent for crowd counting than other types of detectors.

\section{Discussions}
\label{section:Discussions}

In addition to the above results, we also observed the limitations of some detectors and potential improvement directions in the experiments. In this section we will further elaborate on them.

\subsection{Extremely dense scenes and low resolution}

While traditional detectors exhibit commendable performance on prevalent datasets, their efficacy diminishes notably when applied to the ShanghaiTech(A) and UCF\_CC\_50 datasets with pseudo-labels. In scenarios characterized by exceedingly dense crowds, the reduction in the size of pedestrians' head, coupled with diminished distinctive features, adversely impacts the detector's perceptual capabilities. Consequently, in environments with exceptionally high crowd density and limited resolution, we posit that methods grounded in density estimation offer more apt solutions. Apart from that, the detection based methods can still cover most scenarios with the improvement of imaging resolution.

\subsection{Perception of small targets}
In practical scenarios, detecting smaller targets can be a formidable challenge, especially in instances where the scale variance among targets is substantial. This inherent difficulty affects the accuracy of crowd counting methods. Therefore, it is imperative to improve the detection of smaller targets. In Section \ref{Ablation study}, our investigation endeavors to enhance the saliency of small targets within the loss function by amplifying their influence on the heatmap response. This augmentation can mitigate the occurrence of small target omissions.

\subsection{Limitations on two-stage detectors}
Concerning the GTA\_Head dataset, the deployment of two-stage detectors was deliberately omitted due to their observed inefficacy in large-scale target detection experiments. Notably, conventional two-stage detectors, such as Faster R-CNN, necessitate a substantial augmentation in the quantity of proposals generated for potential targets. This augmentation, however, results in a notable diminution of both training and detection efficiency. Furthermore, such an approach imposes a considerable demand on GPU resources, exacerbating the computational burden associated in the detection process.

\subsection{Advantages of anchor free detectors}
According to Table \ref{table8}, the performance of the anchor-based models is worse than that of the other three anchor-free models. Our analysis posits that the efficacy of manually-set anchors featuring fixed dimensions diminishes in the face of extensive scale variations inherent in the task of detecting densely distributed targets. Consequently, we advocate that anchor-free models relying on object center localization exhibit heightened appropriateness for the domain of object detection within scenarios characterized by high-density and intricate spatial configurations.

\section{Conclusion}
\label{section:conclusion}


In this paper, we achieve notable advances in the field of crowd counting with head detection networks, including two pivotal contributions. Firstly, through the introduction of the Mask Focal Loss, we achieved substantial improvements in detector performance. This novel loss function, characterized by the re-division and re-weighting of positive and negative samples, facilitated enhanced perception of the head region and accurate heatmap prediction. Moreover, the Mask Focal Loss demonstrated its versatility by serving as a comprehensive framework for classification losses based on both heatmap and binary feature maps. Besides, we also propose the GTA\_Head dataset, marked by meticulously annotated head bounding boxes. GTA\_Head, featuring diverse and complex scenes with multiple density distributions, serves as a crucial resource for training and evaluating crowd counting networks. Extensive experiments on multiple datasets demonstrate the effectiveness of our proposed approach, as evidenced by reduction of MAE and RMSE by up to 47.03\% and 61.99\%.

Our findings revealed that in high-density scenarios, anchor-free detectors, when coupled with Mask Focal Loss, outperformed anchor-based detectors. Consequently, we assert that the anchor-free model presents superior competence in addressing the challenges posed by crowd counting tasks. For future work, it is also promising to explore new network structures for learning more representative crowd features, especially in the highly-dense and small target cases. For new applications, we believe that our method of crowd counting can not only be used in the security field, but also be extended to the detection and counting of other types of objects, such as cell counting in biomedical engineering, cancer diagnosis, vegetable and fruit or sheep counting in smart agriculture. Improving the detection accuracy of the model in these fileds through our proposed loss function.


\section*{Acknowledgment}

This work was supported in part by Grants of National Key R\&D Program of China 2020AAA0108304, in part by the National Nature Science Foundation of China under Grant 62171288, and in part by the Shenzhen Science and Technology Program under Grant JCYJ20190808143415801. The authors would also like to thank the anonymous reviewers for useful and insightful comments.

\section*{Authors contribution statement}
X.Z., G.W. and W.L. conceived the concepts, advised on the design and implementation of the experiments, conducted experiments, analyzed the data and wrote the manuscript. Z.W. and Y.D. coordinated and supervised the research. All authors read, edited, and discussed the manuscript and agree with the claims made in this work. 
\section*{Ethical and informed consent for data used}
Ethical approval was not sought for the present study because this article does not contain any studies with human or animal subjects.
\section*{Conflict of interest statement}
The authors declare that there is no conflict of interests regarding the publication of this article
\section*{Data availability statement}
The data that support the findings of this manuscript are openly available in GTA\_Head (https://github.com/gkw0010/GTAV Head-dataset.)


\bibliography{sn-bibliography}

\end{document}